\documentclass[a4paper,11pt]{article}
\usepackage{graphicx} 
\usepackage{xspace}
\def\etc{{\em etc.}\@\xspace}
\def\eg{{\em e.g.}\@\xspace}
\def\ie{{\em i.e.}\@\xspace}

\usepackage[T1]{fontenc}  
\usepackage{amsmath}
\usepackage{array}
\usepackage{url} 
\usepackage[utf8]{inputenc}

\usepackage{titling}

\pretitle{\begin{flushleft}\LARGE\bfseries}
\posttitle{\par\end{flushleft}}
\preauthor{\begin{flushleft}\large}
\postauthor{\par\end{flushleft}}
\predate{\begin{flushleft}}
\postdate{\par\end{flushleft}\vskip 0.5em}

\begin{document}
\title{Multi-Level Sensor Fusion with Deep Learning}

\author{Valentin~Vielzeuf$^{1,2}$, Alexis~Lechervy$^{2}$, Stéphane~Pateux$^{1}$, and~Frédéric~Jurie$^{2}$\\
$^{1}$ Orange Labs, Rennes\\
$^{2}$ Normandie Univ., UNICAEN, ENSICAEN, CNRS\\
 }
\maketitle

\begin{abstract}
In the context of deep learning, this article presents an original deep network, namely CentralNet, for the fusion of information coming from different sensors. This approach is designed to efficiently and automatically balance the trade-off between early and late fusion (\ie  between the fusion of low-level vs high-level information).  More specifically, at each level of abstraction --~the different levels of deep networks~-- unimodal representations of the data are fed to a central neural network which combines them into a common embedding. In addition, a multi-objective regularization is also introduced, helping to both optimize the central network and the unimodal networks. Experiments on four multimodal datasets not only show state-of-the-art performance, but also demonstrate that CentralNet can actually choose the best possible fusion strategy for a given problem.

\end{abstract}

\section{Introduction and Related Work}
In recent years, the multiplication of sensor technologies has resulted in a massive increase of multimodal data such as videos, RGB+D data, \etc.  The fusion of  modalities has been successful in various domains, from medical imaging to autonomous driving, including natural language processing or image analysis. The main motivation behind the use of multimodal data is to combine complementary information from different modalities, leading to better decisions than using only one. Many problems that benefit from multiple modalities can be found in the recent deep learning literature such as video classification~\cite{wang2017truly}, emotion recognition~\cite{ringeval2017summary}, or human activity recognition~\cite{baradel2018glimpse}.

A first basic taxonomy of multimodal fusion~\cite{lahat2015multimodal} distinguishes between the approaches by the level of  representation at which the fusion is done, as deep neural networks process the data at different levels of abstraction, the different layers of the networks. A universal optimal level of fusion (early or late) does not exist, as it is task dependent. For example, Simonyan~\textit{et al.}~\cite{simonyan2014two}'s two stream convolutional neural network for human activity recognition is fusing modalities at prediction level (late fusion). In the same way, in audiovisual emotion recognition, better performance is observed using late fusion \cite{kim2017multi,vielzeuf2017temporal}. In contrast, Arevalo~\textit{et al.}~\cite{arevalo2017gated} propose a Gated Multimodal Unit, weighting the modalities features using relatively early fusion information on a textual-visual dataset. Finally, Chen~\textit{et al.}~\cite{chen2017multimodal} follow an early fusion hard-gated approach for textual-visual sentiment analysis. 

Nevertheless, the early versus late fusion paradigm can be insufficient to describe some approaches. Indeed, Neverova~\textit{et al.}~\cite{neverova2014multi} highlight that there are benefits to fuse similar modalities earlier. 
Another example is the multilayer approach of Yang~\textit{et al.}~\cite{yang2016multilayer} performing the fusion on human activity videos with a boosting algorithm applied to all layers. Kang~\textit{et al.}~\cite{kang2017contextual} also use a multilayer framework where the layers' representations are aggregated into a unique representation. 

Finally, it is worth mentioning some approaches relying on regularization: Andrew~\textit{et al.}~ \cite{andrew2013deep} propose the deep Canonical Correlation Analysis method aiming at maximizing the correlation between the multimodal representations, while Neverova~\textit{et al.}~ \cite{neverova2016moddrop,li2016modout} propose modDrop and modout regularizations, consisting in dropping modalities during training. 

The proposed method borrows from both visions: it is a multilayer framework and uses a constrained multi-objective loss. It builds on existing deep convolutional neural networks designed to process each modality independently. We suggest connecting these networks, at different levels, using an additional central network dedicated to the projection of the features coming from the different modalities into the same common space. In addition, the global loss allows to back propagate some global constraints on each modality, coordinating their representations. The proposed approach aims at automatically identifying an optimal fusion strategy for a given task. The approach is multi-objective in the sense that it simultaneously tries to minimize per modality losses as well as the global loss defined on the joint space. 
This article is an extension of \cite{vielzeuf2018centralnet}. 

After presenting our approach in the next section, Section~\ref{Experiments} gives an extended experimental validation of the approach showing the advantage of our method on 4 different multimodal tasks.

\section{CentralNet}
\label{Methods}
In this text we use the phrase \textit{multimodal fusion} to refer to the combination of information provided by different media, under the form of their associated features or the intermediate decisions. More formally, if  $M^1$ and $M^2$ denote the media and $D^1$ and $D^2$ the decisions inferred respectively from $M^1$ and $M^2$, the goal is to make a better decision $D^{1,2}$ using both $M^1$ and $M^2$. More than 2 modalities can be used. This paper addresses the case of classification tasks, but any other task, \eg regression, can be addressed in the same way.  

\begin{figure}[tb]
\begin{minipage}{0.48\linewidth}
\centering
\includegraphics[width=\linewidth]{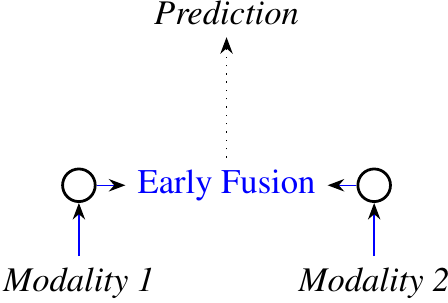}\\
(a)
\end{minipage}
\begin{minipage}{0.48\linewidth}
\centering
\includegraphics[width=\linewidth]{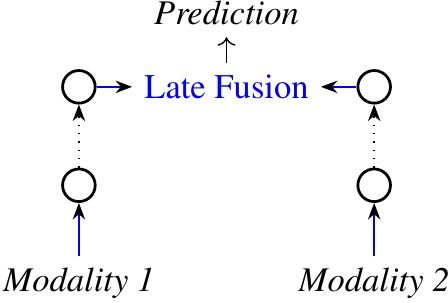}\\
(b)
\end{minipage}
\begin{minipage}{\linewidth}
\centering
\includegraphics[width=\linewidth]{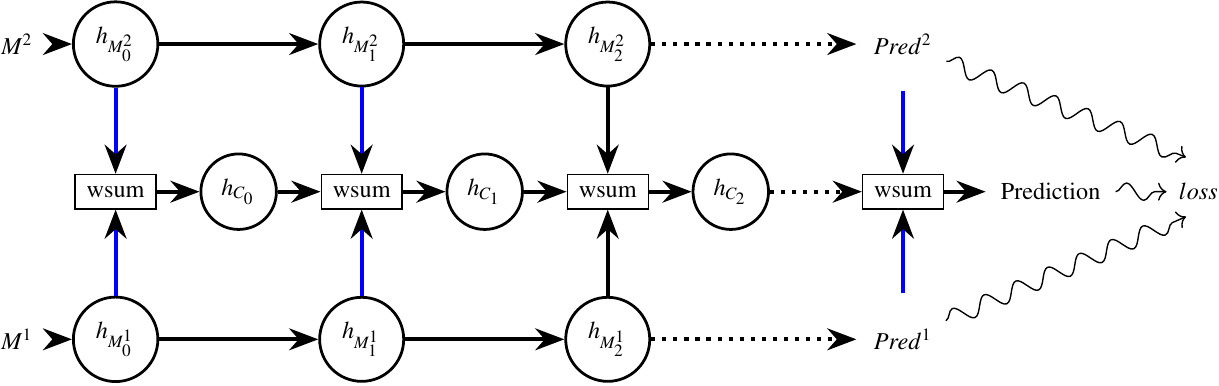}\\
(c)
\end{minipage}
\caption{{(a) Early fusion method}, fusing the low-level representations of the modalities.
(b) {Late fusion method}, fusing the high-level representations of the modalities (or even predictions) obtained for each unimodal neural network.
{(c) Our CentralNet fusion model}, using both unimodal hidden representations and a central joint representation at each layer. The fusion of the unimodal representations is done in this case using a learned weighted sum.
For the sake of simplicity, only the overall synoptic views of the architectures are represented. More details are provided in Section~\ref{Methods}.
\label{schemaCentral}}
\end{figure}

This article focuses on the case of neural nets, for which the data is sequentially processed by a succession of layers. We assume having one neural net per modality, capable of inferring a decision from each modality taken in isolation, and want to combine them. One recurrent question with multimodal fusion is where the fusion has to be done: close to the data (early fusion as in Fig.\ref{schemaCentral}.a), at the decision level (late fusion as in Fig.\ref{schemaCentral}.b) or in between. In the case of neural networks, fusion can be performed at any level between the input and the output of the different unimodal networks. 

For simplicity, we assume that the extracted features (at the input of the fusion layers) have the same dimensionality. If it is not the case, the features can be projected, \eg with 1x1 convolutional layers or zero padded to give them the same size. In practice, the last convolution layers or the first dense layers of separately trained unimodal networks can be used as features.  

\paragraph{CentralNet Architecture}
The architecture of CentralNet~\cite{vielzeuf2018centralnet} is a neural network which combines the features issued from different modalities, by taking as input of each one of its layers, a weighted sum of the layers of the corresponding unimodal networks and of its own previous layers. This is illustrated in Figure~\ref{schemaCentral}(c). Such fusion layers can be defined by the following equation:
\begin{equation} h_{C_{i+1}} =  \alpha_{C_{i}} h_{C_{i}} + \sum_{k=1}^{n} \alpha_{M_{i}^{k}} h_{M_{i}^{k}} 
\label{mainEqu}
\end{equation}
where $n$ is the number of modalities, $\alpha$ are trainable scalar weights, $h_{M_{i}^{k}}$ is the hidden representation of each modality at layer $i$, and $h_{C_{i}}$ is the central hidden representation.
The resulting representation $h_{C_{i+1}}$ is then fed to an operating layer cell (which can be a convolutional or a dense layer followed by an activation function).  

Regarding the first layer of the central network ($i=0$), as we do not have any previous central hidden representation, we only weight and sum the representations of all modalities ($M^i$) issued from unimodal networks. 
At the output level, the last weighted sum is done between the unimodal predictions and the central prediction. Then, the output of the central net is used as the final prediction.

\begin{figure}[tb]
\includegraphics[width=.5\linewidth]{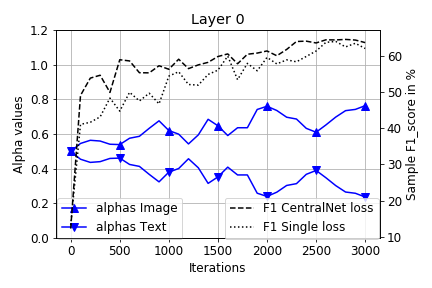}
\includegraphics[width=.5\linewidth]{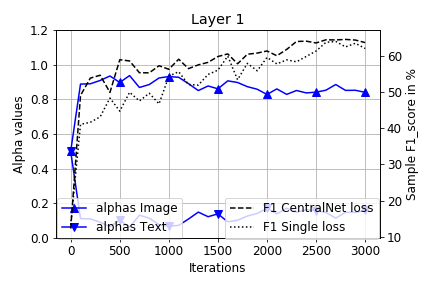}
\includegraphics[width=.5\linewidth]{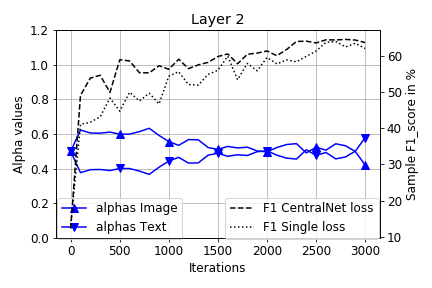}
\includegraphics[width=.5\linewidth]{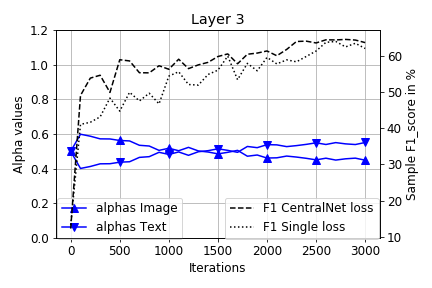}
\caption{Evolution of $\alpha$ values (normalized to 1) and F1-scores (on the test set) during training on MM-IMDb}
    \label{weights_during_training}
\end{figure}

\paragraph{Training the CentralNet model}
All trainable weights of the unimodal networks, the ones of the CentralNet and the fusion parameters $\alpha_{M_{i}^{k}}$, are optimized jointly by applying a stochastic gradient descent using the Adam approach.
The global loss is defined as: $ loss = loss_{C} +  \sum_{k=1}^n loss_{M^{k}} $ where $loss_{C}$ is the (classification) loss computed from the output of the central model and $loss_{M^{k}}$ the (classification) loss when using only modality $k$. As already observed by Neverova~\textit{et al.}~\cite{neverova2016moddrop}, when dealing with multimodal fusion, it is crucial to maintain the performance of each unimodal neural networks. It is the reason why the global loss also includes unimodal losses. It helps generalization by acting as a multi-objective regularization (see Fig.~\ref{weights_during_training}).

Overall, CentralNet is easy to implement and can be built on top of existing architectures known to be efficient for each modality. We refer to \cite{vielzeuf2018centralnet} for implementation details. The resulting values of the $\alpha$ weights allow some interesting interpretations on where modalities are combined. For instance, getting $\alpha_{M_{i}^{k}}$ values close to 0 for $i > 0$ is equivalent to early fusion, while having all the $ \alpha_{C_{i}}$ close to 0 up to the last weighted sum would be equivalent to late fusion.

\section{Experiments}
\label{Experiments}
This section provides  extensive evaluations of CentralNet on four multimodal datasets, namely Montalbano~\cite{escalera2014chalearn}, MM-IMDb~\cite{arevalo2017gated}, the 'animal' subset of Audioset~\cite{gemmeke2017audio} and AFEW~\cite{dhall2012collecting}. 
For each dataset, CentralNet is compared with four different fusion approaches: \textit{early} fusion (concatenation of low-level features), \textit{late} fusion (concatenation of the unimodal scores), \textit{ModDrop} (our own implementation of ModDrop, giving the same results as the one in Neverova~\textit{et al.}~\cite{neverova2016moddrop}) and \textit{Gated Multimodal Unit} (GMU) (re-implemented following Arevalo \textit{et al.}~\cite{arevalo2017gated}, our own implementation reaching slightly better performance than the original one). To assess the statistical significance of the results, 64 models with different random initialization are evaluated and we report the average performance and the standard deviation.

\begin{figure}[tb]
    \centering
    \includegraphics[width=\linewidth]{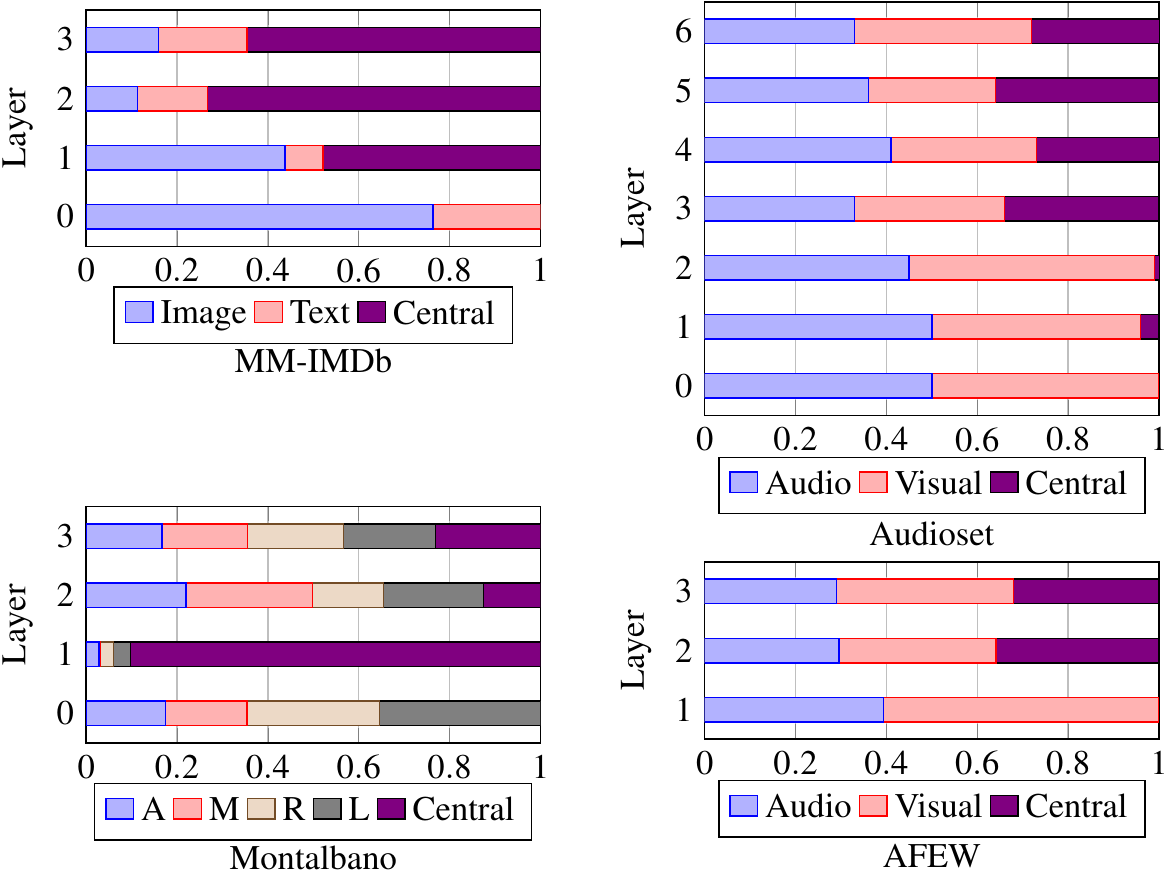}
        \caption{Visualization of the $\alpha_i$ weights after training. They are displayed as the percentage given to each modality and to the central hidden representations across layers and datasets. We observe that the learned fusion strategy is different for each dataset.}
    \label{all_weights}
\end{figure}

\begin{table}[tb]
\centering
\caption{Overall performance on the different datasets.}
\label{all_results}
\centering
\resizebox{\columnwidth}{!}{\begin{tabular}{lccccc}
\hline
                    & Montalbano~\cite{escalera2014chalearn}     & MM-IMDb~\cite{arevalo2017gated}        & AFEW~\cite{dhall2012collecting}         & Animals~\cite{gemmeke2017audio}  \\
Fusion meth. & Macro accuracy & F1-score & Accuracy & F1-score \\ \hline
Best moda.      & 88.00$\pm 0.30$           & 60.20$\pm 0.2$          & 52.5$\pm 0.7$            & 55.2        \\
Early               & 97.81$\pm 0.05$          & 63.0$\pm 0.24$          & 40.1$\pm 1.2$          & 50.1    \\
Late                & 97.54$\pm 0.02$          & 63.5$\pm 0.14$          & 55.3$\pm 1.2$          & 58.3    \\
ModDrop ~\cite{neverova2016moddrop}            & 98.19$\pm 0.03$          & 62.4$\pm 0.15$          & 53.2$\pm 0.9$          &         \\
GMU~\cite{arevalo2017gated}                 & 97.98$\pm 0.04$          & 63.0$\pm 0.17$          & 55.1$\pm 0.8$          & 58.7      \\
\textbf{CentralNet} & \textbf{98.27$\pm$ 0.03} & \textbf{63.9$\pm$ 0.12} & \textbf{57.2$\pm$ 0.4} & \textbf{59.1} & \textbf{} \\ \hline
\end{tabular}}
\end{table}

\subsection{Experiments on Montalbano}
Montalbano~\cite{escalera2014chalearn} gathers more than 14,000 samples of 20 Italian sign gesture categories. These videos were recorded with a Kinect sensor, capturing audio, skeleton joints, RGB and depth (4 modalities). The task is to recognize the gestures from the video data. The performance is measured as the macro accuracy, which is the average of the accuracy for each class .
We used the features given by \cite{neverova2016moddrop}: audio features (size 350), motion capture of the skeleton (size 350), RGB+depth left/right-hand features (size 400). Features are zero-padded to give vectors of size 400, which is possible as done after convolution layers. Fusion architecture includes 1 multilayer perceptron per modality, each having 3 layers of size: $400\times 128$,  $128\times 42$,  $42\times 21$. CentralNet  connects the 3 layers of the different modalities to the central network.

We use dropout (50\% dropping) and batch normalization. The learning rate starts at 0.05 and is multiplied by 0.96 at each epoch, the batch size is of 42, containing two samples of each class, and the model is trained on 100 epochs for all experiments. For ModDrop, the modality drop probability is of 0.5. 

The performance obtained with each modality varies from 46\% (left hand) to 88\% (motion capture of skeletons). The first column of Table~\ref{all_results} shows that both late fusion and early fusion give significant improvement compared to the best modality, thus suggesting some complementarity between the modalities. 
CentralNet outperforms other approaches. 
In addition, Fig.~\ref{all_weights}, gives more information regarding the learned fusion strategy by visualizing the weights proportion at each layer. At the first layer (layer 0), the weights reflect the dimensionality of the modalities. At the next layer, almost no information is taken from the modalities, while at layers 2 and 3, the weight given to each modality and to the central representation are relatively similar. This may be interpreted as a hybrid fusion strategy, mixing "early" and "late" fusions.

\subsection{Experiments on MM-IMDb}
The MM-IMDb dataset~\cite{arevalo2017gated} contains 25,959 samples. One sample is composed of a movie plot, its associated poster and its genres (e.g. Drama, Fantasy). The task is to predict a movie's genre based on its plot and on its poster (2 modalities). One movie can belong to more than one of the 23 possible genres. The task hence has to be evaluated as a multilabel classification task. As in~\cite{arevalo2017gated,kiela2018efficient}, we measure performance with the micro F1 score. More details about other F1 score measures may be found in our workshop paper~\cite{vielzeuf2018centralnet}. 
For these experiments, we use the features kindly provided by the authors of~\cite{arevalo2017gated}. The visual features of size 4096 are extracted from the posters using the VGG-16~\cite{simonyan2014very} network pre-trained on Imagenet. The 300-d textual ones are computed with a fine-tuned word2vec~\cite{mikolov2013distributed} encoder.

We build a multilayer perceptron on top of the features of each modality. For both modalities, the network has 3 layers of size $\textit{input$\_$size}\times 4096$, $4096\times 512$ and $512\times 23$. The CentralNet architecture is the same, taking 4096-d vectors as inputs, zero-padding the textual features to reach the visual features' size.

We use dropout (50\% dropping) and batch normalization. The learning rate is 0.01 and the batch size is 128. For Moddrop, the modality drop probability is of 0.25. The loss of the models is a weighted cross entropy, with a weight of 2.0 on the positive terms to balance precision and recall. More formally, the loss is:
\begin{equation}
    loss = -y\log(2\sigma(pred)) - (1 - y) \log(1 - \sigma(pred))
\end{equation}
with $\sigma(pred)$ the sigmoid activation of the last output of the network and $y$ the multiclass label.
As recommended by Arevalo~\textit{et al.}~\cite{arevalo2017gated}, we also use early stopping on the validation set.

Table~\ref{all_results} reports the performance measured on the test set. First of all, the worst std. dev. we observe is very small.
Second, the best modality is text (60.2) and it outperforms image (47.8) by a large margin. Third, we note that the early and late fusion methods bring a large improvement and are not far from the CentralNet, which gives the best performance. Fig.~\ref{all_weights} shows that CentralNet gives more weight to the first layers, indicating that an "earlier fusion" strategy is privileged in this case, even if the two modalities contribute significantly at all levels.  Fig.~\ref{weights_during_training} shows the evolution of the (normalised to 1) $\alpha$s during training. We can see that the $\alpha$s converge to stable values, favoring images at the first 2 layers (early fusion) and slightly favoring text in the last 2 layers (late fusion). We also see that the CentralNet loss makes the convergence faster in addition to making the performance slightly better.

\subsection{Experiments on AFEW}
AFEW~\cite{dhall2012collecting} is a dataset of 1,156 video clips extracted from movies recorded by a standard video-camera. The task is to predict the emotion of a person, among seven classes. The performance is measured as the average of the accuracy on 64 runs (and standard deviation is also provided).

The audio modality is encoded as 1,583-d feature vector, obtained with the OpenSmile toolkit, while the visual modality is following the same approach as~\cite{vielzeuf2018occam} and is a 512-d feature vector. 
CentralNet is made of a multi-layer perceptron with one hidden layer [1583 -> 128 -> 7]. To perform the first weighted sum, the visual modality is zero-padded to the size of the audio modality.
The choice of this architecture with a small number of parameters can be explained by the great risk of over-fitting on such a small dataset.

The third column of Table~\ref{all_results} reports the performance measured on the validation set during the different experiments, as the test set is not available. The visual modality achieves an accuracy of 52.5\%, while the audio accuracy is of 34.2\%. Thus, the visual modality is performing best by a large margin. Early fusion brings no benefits compared to the visual modality, while all other methods are better and more stable (smaller standard deviations). CentralNet brings the largest improvement and prefers a relatively late fusion strategy, as shown in Fig.~\ref{all_weights} where the central weights are small in the first layers.

\subsection{Experiments on Audioset}
Audioset~\cite{gemmeke2017audio} is a large audiovisual dataset containing YouTube videos (therefore recorded by different types of video-cameras). We only focus on a subset containing 19,842 videos of 10 seconds of different pets (named "domestic animals, pets" in the ontology of audio set). These videos are annotated with 211 different class labels, which are not only relative to animals but also to context (e.g. "ringtone", "meow"). Therefore, the classification task is multilabel. To take this issue into account, we propose to use the micro F1-score.

Modalities are a spectrogram of the audio data, resized to 512x512 and the video sequence subsampled to 16x256x256 images. 
An inflated Resnet-50 is applied on the image sequence, in the same fashion as in~\cite{baradel2018glimpse}. For audio inputs, we use a model containing six convolutional layers (kernel size is 4 and stride is 2), an average two-dimensional pooling layer (7 by 7) and a fully connected layer. The first convolutional layer is outputting 64 filters, and the number of filters is doubled at each layer until having 2048 filters. The associated 8x8 maps are pooled to 1x1 and the obtained 2048 features are fed to the fully connected layer. Between each layer, we use batch normalization and reLU activation.

The CentralNet architecture is identical to the audio CNN with the first two convolutional layers removed (beginning with 256 filters). The weighted sum of the features, at a given layer, is the average of the features of the inflated Resnet-50 alongside the temporal dimension. 

As training is time consuming, we made only one run. The 4th column of Table~\ref{all_results} reports the performance measured on this dataset. ModDrop score is not provided as the authors were not able to make it work on this dataset (convergence issues). Audio and visual models alone respectively achieve F1-scores of 55.2\% and 54.8\%. Late fusion is outperformed by GMU,  itself outperformed by CentralNet. We note that  early fusion did not bring any improvements here, mainly because the combined low-level features may not be relevant to each other. This is in line with the Fig.~\ref{all_weights}, where we observe that the central contribution is ignored in low-level layers, and progressively taken into account when getting closer to prediction level.  

\subsection{Comparison with other fusion methods}
Table \ref{all_results} shows that the proposed method outperforms 5 alternative fusion models (including the very recent approaches of \cite{neverova2016moddrop} and \cite{arevalo2017gated}) on 4 different datasets. The difference is statically significant according the two sample z test, with z values between 12 and 47 for 64 runs. Additional computational cost is marginal, as the number of parameters of the central network is marginal compared to the number of parameters of the unimodal networks, and does not require any additional iterations at training time.

\section{Conclusions}
This article, presenting CentralNet and experimentally validating it on 4 datasets, consolidates the idea that the best fusion strategy heavily depends on datasets and tasks. It is underlined by the fact that CentralNet is performing best on all of the 4 datasets. This is explainable by its capacity to automatically find at which layers the multimodal fusion should be performed. Moreover, the interpretation of the different fusion strategies is in line with the observations made in the literature on optimal depths for fusion. Future works will explore more flexible frameworks, using other operations than weighting sums and allowing more than one by one layer combinations.
 
\normalsize

\bibliographystyle{plain}
\bibliography{egbib}

% that's all folks
\end{document}